\documentclass[10pt,journal]{IEEEtran}

\usepackage{amsmath}
\usepackage{graphicx}
\usepackage{cite}
\usepackage{times}
\usepackage{epsfig}
\usepackage{amsmath}
\usepackage{amssymb}
\usepackage{booktabs}
\usepackage{multirow}
\usepackage[pagebackref=true,breaklinks=true,letterpaper=true,colorlinks,bookmarks=false]{hyperref}
\usepackage{color}
\usepackage{algorithm}
\usepackage{algpseudocode}
\usepackage{pifont}
\usepackage{authblk}

\makeatletter
\newcommand*{\rom}[1]{\expandafter\@slowromancap\romannumeral #1@}
\makeatother

\begin{document}

\author{Abrar~H.~Abdulnabi,~\IEEEmembership{Student~Member,~IEEE},
	 Bing~Shuai,~\IEEEmembership{Student~Member,~IEEE},
	 Zhen~Zuo,~\IEEEmembership{Student~Member,~IEEE},
	 Lap-Pui~Chau,~\IEEEmembership{Fellow,~IEEE},
	 and~Gang~Wang,~\IEEEmembership{Senior~Member,~IEEE}
	\thanks{A. H. Abdulnabi is working with both the Rapid-Rich Object Search (ROSE) Lab at Nanyang Technological University, Singapore, and the Advanced Digital Sciences Center (ADSC), Illinois at Singapore Pt Ltd, Singapore, Email: abrarham001@ntu.edu.sg. B. Shuai is with ROSE, Z. Zuo is with ROSE, L. Chau and G. Wang are with the Department of Electrical and Electronic Engineering, Nanyang Technological University, Singapore, Emails: elpchau@ntu.edu.sg and wanggang@ntu.edu.sg respectively. Address of ADSC: Advanced Digital Sciences Center, 1 Fusionopolis Way, Illinois at Singapore, Singapore 138632. Address of ROSE: The Rapid-Rich Object Search Lab, School of Electrical and Electronic Engineering, Nanyang Technological University, Singapore, 637553.}
}

\title{Multimodal Recurrent Neural Networks with Information Transfer Layers for Indoor Scene Labeling}

\markboth{}
{Shell \MakeLowercase{\textit{Abrar et al.}}}

\maketitle

\begin{abstract}
This paper proposes a new method called Multimodal RNNs for RGB-D scene semantic segmentation. It is optimized to classify image pixels given two input sources: RGB color channels and Depth maps. It simultaneously performs training of two recurrent neural networks (RNNs) that are crossly connected through information transfer layers, which are learnt to adaptively extract relevant cross-modality features. Each RNN model learns its representations from its own previous hidden states and transferred patterns from the other RNN’s previous hidden states; thus, both model-specific and cross-modality features are retained. We exploit the structure of quad-directional 2D-RNNs to model the short and long range contextual information in the 2D input image. We carefully designed various baselines to efficiently examine our proposed model structure. We test our Multimodal RNNs method on popular RGB-D benchmarks and show how it outperforms previous methods significantly and achieves competitive results with other state-of-the-art works.
\end{abstract}


\IEEEpeerreviewmaketitle

\section{Introduction}
\IEEEPARstart{D}{ata} comes from different sources and in different forms; images, videos, texts and audios. Each of which may complement the other in information content. Thus, multiple data modalities are usually more informative for a task than a single data modality. With the enormous availability of various electronic and digital multimedia devices, huge volumes of multimodal data contents are being generated on the Internet daily. However, for real-world applications, these modalities should be first well integrated and appropriately fused to have more comprehensive information contents. Many methods have been developed in multimodal learning to exploit both the different characteristics as well as the shared relationships between different data modalities in order to perform various tasks \cite{hashTMM5,faceTMM2,anranWangTMM1,retrievalTMM3,poseTMM4,me3}. One of the rigid milestones in developing many visual-based data applications is scene understanding. It is mainly applied to understand the contents of an image or video prior performing the target task (e.g. large-scale video retrieval \cite{retrievalTMM3}). Imperatively, scene labeling (i.e; semantic segmentation or image parsing) is a crucial part of understanding an outdoor or indoor captured scene image. The task here is to classify each pixel into its semantic category (belonging to some object or stuff) in an input scene image. 

In our paper, we tackle the problem of RGB-D indoor scene labeling where we process two different data modalities; RGB color channels and Depth planes. Indoor RGB-D scene labeling is one of the most challenging visual classification problems ~\cite{Silberman:indoorSegmentationv1,Silberman:indoorSegmentationv2}. Many applications are built on understanding the surrounding scenes, e.g. social behavior understanding \cite{app2} and objects detection and recognition \cite{app1}.

\begin{figure}
	\centering
	\begin{center}
		\includegraphics[width=0.45\textwidth]{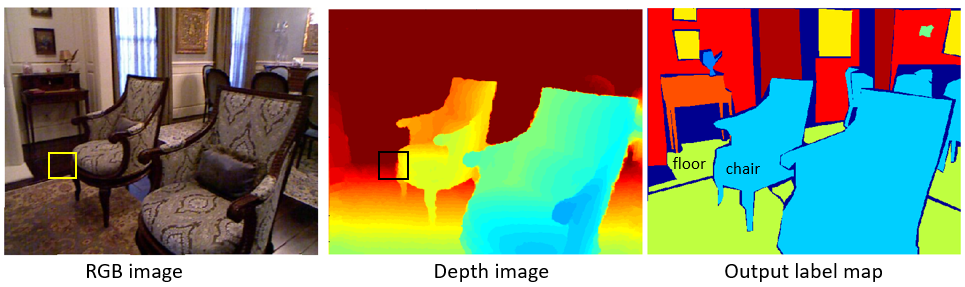}
		\caption {An example scene image that have both the RGB image and the Depth plane. The corresponding Depth information can be utilized to better differentiate ambiguous appearance in the RGB image. \label{Fig_rgbd_5}}
	\end{center}
\end{figure}

This problem is usually addressed as a multimodal learning problem where the task is to exploit and fuse both RGB and Depth modalities to better label each pixel. Depth planes provide informative representation where the RGB representations are ambiguous. For example, Figure \ref{Fig_rgbd_5} show how the depth information can help to distinguish some similar appearance locations in the RGB image. 

Scene labeling in general is a challenging classification problem since the scene image tends to contain multiple messy objects. These objects may also have variations due to factors affecting their appearance and geometry in the image. One key useful strategy is to leverage the neighborhood/contextual information for each pixel within each modality\cite{reviewer21,reviewer22,me1,me2}. Typically, the feature representation of a pixel is extracted from a local patch (cropped from the scene image) containing that target pixel and used for classification. Long-range/global contextual information (distant image patches) are important as well for local pixel classification. However, both local and global contextual information should be utilized adequately to maintain a good balance between the discriminative features and the abstract/top-level features of that pixel feature representation. 

Recently, Recurrent Neural Networks (RNN) has been shown to be very successful on encoding contextual information into local feature representations \cite{ZhouZhen,ShuaiBing,Byeon_2015_CVPR}. Recurrent models have feedback connections so that the current state is engaged in the calculation of the future state. Thus, RNN is effectively used in tasks which require modeling the long and short-range dependency within the input sequence, e.g. speech recognition and natural language processing \cite{Graves_offlinehandwriting,irsoy-drnt,speechRNNGraves,multidiminRNN2DAlexGraves}. We use RNNs to model the contextual information within each modality. However, traditionally, RNN is only used for a single modality signals. In this paper, we introduce a new multimodal RNNs method which models contextual information into local representations from multimodal RGB-D data simultaneously. In our work, we first train Convolution Neural Networks \cite{NIPS2012} (CNNs) to extract features from local RGB-D image patches (from both the RGB images and the Depth planes). These convolutional local features form the input to our multimodal RNNs to further contextualize them and select informative patterns across the modalities. Our model can be easily extended to perform prediction tasks considering more modalities ($>2$).

Our new multimodal RNNs method is built based on the basic quad-directional 2D-RNNs structures \cite{multidiminRNN2DAlexGraves,ShuaiBing}. The quad-directional 2D-RNN contains four hidden states where each is dedicated to traverse the image in specific 2D direction out of four possible directions \textbf{(top-left, top-right, bottom-left and bottom-right)}. To process two modalities which are the RGB image and the Depth plane; our model has two RNNs, one of which is assigned to learn the representations of a single input modality. To connect both RNN modalities and allow information fusion, we develop information transfer layers that crossly connecting the RNNs. The transfer layers are responsible about learning to select and transfer the relevant patterns from one modality to the other modality, and vice versa. Concretely, for every patch in the input image, and during the process of learning the RNN hidden representations for one modality, our method does not only encode the contextual information within its own modality, but also learns to encode relevant contextual patterns from the other modality. As a result, our method can learn powerful context-aware and multimodal features for local pixel representations.

Our method is different from existing deep multimodal learning methods \cite{multimodal1,multimodal2,multimodal3,multimodal4}. They usually concatenate inputs at the beginning or concatenate learned middle level features to extract high-level common features as the representations of the multimodal data. These methods mainly focus on discovering common patterns between different modalities. Although common patterns are important to extract, however these methods are prone to miss important modality specific information that are highly discriminative within a single modality, e.g., some texture patterns inside the RGB channels. Concretely, our model retains modality-specific information by assigning an RNN model to learn features from each modality. Our method also allows sharing information between modalities by using the information transfer layers to adaptively transfer relevant cross-modality patterns. 

Our model is trained end-to-end and the transfer layers are learned to extract relevant across-modality information for each patch of the image. We perform experiments on two popular RGB-D benchmarks, the NYU V1 and V2 and achieve comparable performance with other state-of-the-art methods on them. Additionally, the proposed method significantly outperforms its counterparts baselines (e.g. concatenation of the RGB-D data as the input of RNN models). 

The remaining parts of our paper are summarized as follows: We first discuss the related work in Section \rom{2}. Our proposed model alongside the framework details is presented in Section \rom{3}. Experiments on popular RGB-D benchmarks and results are demonstrated in Section \rom{4}. Finally, we conclude the paper in Section \rom{5}.

\section{Related Work}
Because this work is mainly related to RGB-D scene labeling and RNN, we briefly review the most recent literature.
\\
\\
\textbf{Indoor RGB-D Scene Labeling:} Many papers propose different methods  to solve RGB-D scene labeling \cite{Silberman:indoorSegmentationv1,Silberman:indoorSegmentationv2,Gupta:indooeSegmentation,indoorSemanticYannLecun,rgbdSceneFeatures,reviewer1}. The most popular indoor scene datasets are the NYU depth datasets V1 \cite{Silberman:indoorSegmentationv1} and V2 \cite{Silberman:indoorSegmentationv2}. The initial results \cite{Silberman:indoorSegmentationv1} are generated by extracting SIFT features on the RGB color images in addition to depth maps. Their results prove that depth information can refine the prediction performance.

Further improvements are made to the NYU V1 by the work \cite{rgbdSceneFeatures}, where they adapt a framework of kernel descriptors which converts the local similarities (kernels) to patch descriptors. They further use a superpixel Markov Random Field (MRF) and a segmentation tree for contextual modeling. Other works \cite{indoorSemanticYannLecun,predictinDpethCVPR2015} explore depth information through a feature learning approach; \cite{indoorSemanticYannLecun} learns their features using  convolution neural network on four channels; three from the RGB image and the fourth is the Depth image. Wang et al. \cite{WangAnranUnsupervised} adapt an unsupervised feature learning approach by performing the learning and encoding simultaneously to boost the performance. Another interesting work done by Tang et al. \cite{TangHONVgradients} proposes new feature vectors called Histogram of Oriented Normal Vectors (HONV), designed specifically to capture local geometric properties for object recognition with a depth sensor.

Meanwhile some works in RGB image parsing utilize the label dependency modeling approaches, such as graphical models (Conditional Random Fields CRFs \cite{mcrfSceneLabeling}), and other works perform feature learning to generate hierarchical multiscale features that are capable of capturing the input context, which is successfully applied through the aid of deep convolutional neural networks \cite{sceneParsingPurityTrees,recurrentConvScenLabeling}. In our multimodal recurrent neural networks, we can capture short and long range context within and between input modalities.

\begin{figure}
	\centering
	\begin{center}
		\includegraphics[width=0.5\textwidth]{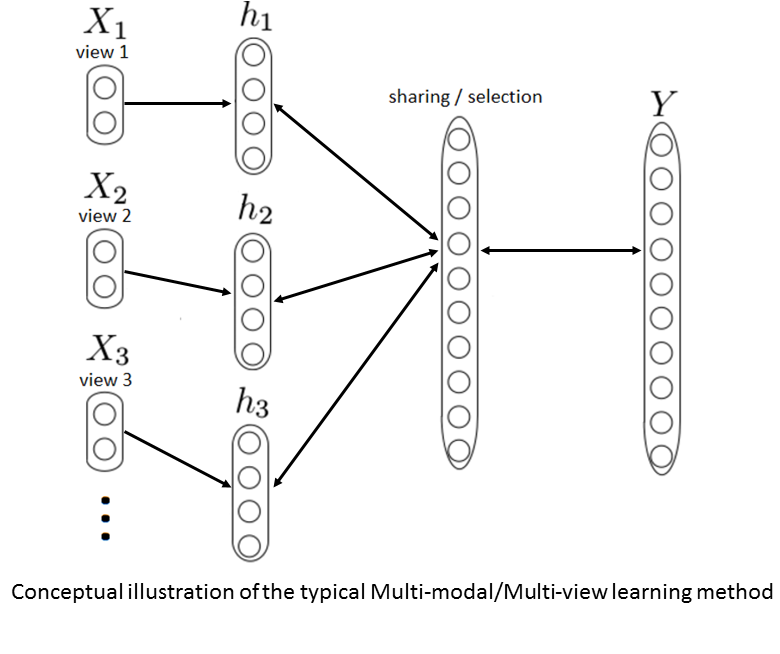}
		\includegraphics[width=0.4\textwidth]{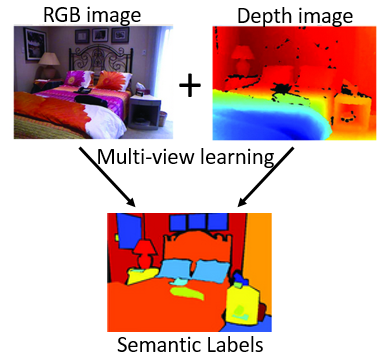}
		\caption {The top part illustrates the concept behind the multi-modal/multi-view learning method where usually different data modalities are combine by various techniques to exploit all information sources. Multi-modal is also efficient to combine and fuse the RGB color channels with Depth planes (bottom part)\label{Fig_rgbd_6}}
	\end{center}
\end{figure}
	
\textbf{Multi-modal Learning:} In contrast to single-view approaches, most multimodal learning methods introduce a separate function to model one data modality and jointly optimize them together \cite{Wang_anew,multiKernalMethod4359380,deepCCA,survayOnMultiview}. Deep networks have been applied to learn features over multiple views \cite{multimodal1,multimodal2,multimodal3,multimodal4}. Multimodal learning improves the classification performance through exploiting the sharable knowledge of the data across modalities. 
	
Figure\ref{Fig_rgbd_6} shows the general method of how multimodal learning is applied. Notice that the key idea here is to allow fusion/sharing between the modalities at some point so that the join space is able to capture the relationships between the input modalities. Typical methods in the literature can be categorized based on whether they combine on feature level \cite{Dalal:2006:HDU:2168483.2168522,depthIntensityPedistrain,Intelligent1603551}, or classifier level \cite{Rohrbach09high-levelfusion,classifierLevelRoboticGrasp}. Some works perform preprocessing (module level) steps on various modalities to generate helpful cues before the learning process, as in \cite{Multi-cue2007,DepthMobile409092,Dynamic3D4270171}. Fusion can be done by combining all features into one high-dimensional vector, or by jointly training multiple classifiers to maximize the mutual agreement on distinct modalities of the input data\cite{TrackingMutiTaskMultiView6751190,GraphMutiTaskMultiView}. These methods employ typical regularizations which are applied to explore shared visual knowledge, such as group structured sparsity \cite{groupStructionRegression}. 

We can also group multimodal methods according to their training procedures into three main categories: Co-training\cite{Wang_anew,survayOnMultiview}, Multiple kernel learning \cite{multiKernalMethod4359380,survayOnMultiview}, and Subspace learning\cite{Hotelling1936,deepCCA,ccaOLD,survayOnMultiview}. Co-training algorithms tend to train alternatingly to maximize the mutual agreement on two distinct views of data. Multiple kernel learning approaches improve the performance by exploiting different types of kernels that correspond naturally to different views, and combine these kernels either linearly or non-linearly. Meanwhile, Subspace learning methods are mainly similar in obtaining a latent subspace that is shared by multiple modalities. Another interesting line of work proposed by Gupta et al. \cite{crossGupta}, where they propose to transfer supervision between images from different modalities. Their model is able to learn representations for unlabeled modalities and can be used as a pre-training procedure for new modalities with limited labeled data. In our work and different from all previous methods, we introduce information transfer layers between two RNN modalities to perform the multimodal learning task simultaneously. 
	
\textbf{Recurrent Neural Networks (RNNs):} A recurrent model refers to a model which has connections between its units to form a directed cycle, for example, when a feedback connection from the current state is engaged in the calculation of the future state. This structure creates a sort of complementary internal state for the network, which then allows it to exhibit dynamic temporal behavior. RNN is effectively used in tasks which require sequence modeling, like speech recognition, handwriting recognition, and other natural language processing tasks \cite{Graves_offlinehandwriting,irsoy-drnt,speechRNNGraves,multidiminRNN2DAlexGraves}. 
	
One major drawback in the standard RNN is the vanishing gradient problem \cite{Hochreiter01gradientflow}. This drawback limits the context range of the input data, because the capacity of the model is limited to capture enough long dependencies. To address this problem, Hochreiter and Schmidhuber \cite{LSTMRNN} propose the Long Short Term Memory (LSTM), where they treat the hidden layer as multiple recurrently connected subnets, known as memory blocks, thus allowing the network to store and access information over long periods of time. Graves et al. extend the idea of unidirectional LSTM network into bidirectional networks which have shown good improvements over the unidirectional networks \cite{BidirectionalLSTMRNN,BidirectionalLSTMRNNImprovememnt}.

\begin{figure}
	\centering
	\begin{center}
		\includegraphics[width=0.4\textwidth]{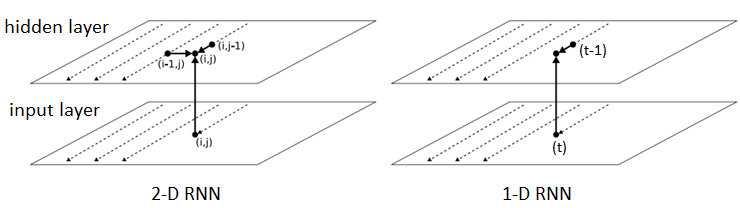}
		\caption {Illustration of the one-dimensional (1D) RNN (right side) and multidimensional (e.g. two-directional) 2D RNN (left side) \cite{multidiminRNN2DAlexGraves}. The key idea is to replace the single recurrent connection in the 1D-RNNs with as many recurrent connections as there are dimensions in the input data.\label{Fig_rnn_2}}
	\end{center}
\end{figure}

Graves et al. also extend the one-dimensional RNN into multidimensional one \cite{multidiminRNN2DAlexGraves,Graves_offlinehandwriting} as shown in Figure\ref{Fig_rnn_2}. The key idea is to replace the single recurrent connection in the 1D-RNNs with as many recurrent connections as there are dimensions in the input data. Another interesting work from Graves et al. investigates deep structures of RNNs \cite{speechRNNGraves}, which is also successfully applied in opinion mining by Irsoy et al. \cite{irsoy-drnt}. In our work we evaluate both deep and LSTM structures alongside the basic quad-directional 2D-RNNs. We found that there is no significant difference in the label prediction performance before and after stacking multiple layers of the network or engaging the LSTM units with the basic quad-directional 2D-RNNs. Thus, we only show the results of our multimodal-RNNs model using the basic quad-directional 2D-RNNs.

\section{Model Framework}
We first extract convolutional features from local RGB-D patches using our trained CNN models. Then, our multimodal-RNNs are developed to further learn context-aware and multimodal features based on the convolutional features. Afterwards, a softmax classifier is trained to classify each patch into its semantic category. Different from traditional single modality RNN, our multimodal-RNNs also have transfer layers to learn to extract relevant contextual information across both modalities at each time step. Below, we first introduce the traditional RNNs.

\textbf{1D-RNN and 2D-RNN:} The popular Elman-type 1D-RNN \cite{1D-RNN-Elman} and its 2D version are designed to capture the dynamic behavior of the signal over time, so that the hidden representations can capture the contextual information from the first time step until the current time step. Its forward pass is formulated as the following:
\begin{equation}
	\small
	\label{1D-RNN-formula}
	\begin{aligned}
		h^{(t)} &= f(Ux^{(t)} + Wh^{(t-1)} + b) & \\ 
		y^{(t)} &= g(Vh^{(t)} + c)
	\end{aligned}
\end{equation}
where $x^{(t)}$, $y^{(t)}$ and $h^{(t)}$ are the input, output and hidden neurons at the time t respectively. The functions $f(.)$ and $g(.)$ are element-wise non-linear functions with bias terms $b$ and $c$, and the matrices $U$, $W$ and $V$ are the input to hidden, hidden to hidden and hidden to output weights, respectively.
The 2D-RNN \cite{multidiminRNN2DAlexGraves,ZhouZhen,ShuaiBing} is generalized from the 1D-RNN so that the data propagation will come from two dimensional neurons instead of one, thus the formulation becomes:
\begin{equation}
	\small
	\label{2D-RNN-formula}
	\begin{aligned}
		h^{(i,j)} &= f(Ux^{(i,j)} + Wh^{(i-1,j)} + Wh^{(i,j-1)} + b) & \\ 
		y^{(i,j)} &= g(Vh^{(i,j)} + c)
	\end{aligned}
\end{equation}
\begin{figure}
	\begin{center}
		\includegraphics[width=0.8\linewidth]{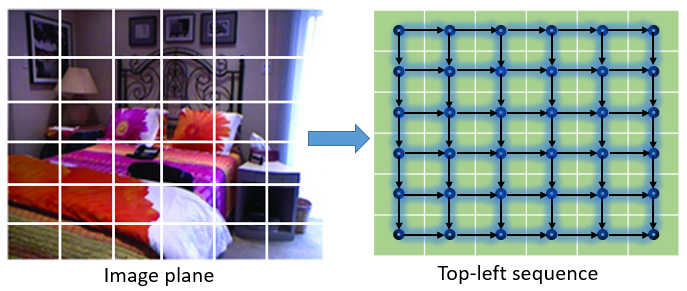}
	\end{center}
	\caption{2D-RNN structure is a generalization of 1D-RNN. A visualization of processing patches in 2D image is shown. The scan is only presented in one direction (top-left). Here, the previous hidden states from the top and the left sided are encoded into the current hidden state, i.e. the previous top and left neighbored hidden state information are engaged in calculation of the current network state alongside the current patch at the same location. An image is approximated through four directions: top-left, bottom-left, top-right and bottom-right accordingly. Thus, we use the quad-directional 2D design to accumulate information from all sides.}
	\label{2d-alex}
\end{figure}
We can notice now the propagation is in a 2D-plane from top left regions and continues to flow until the end of the 2D sequence (in our case the image patch sequence), where $(i,j)$ denotes the location of pixels or patches.

\begin{figure}
	\begin{center}
		\includegraphics[width=1\linewidth]{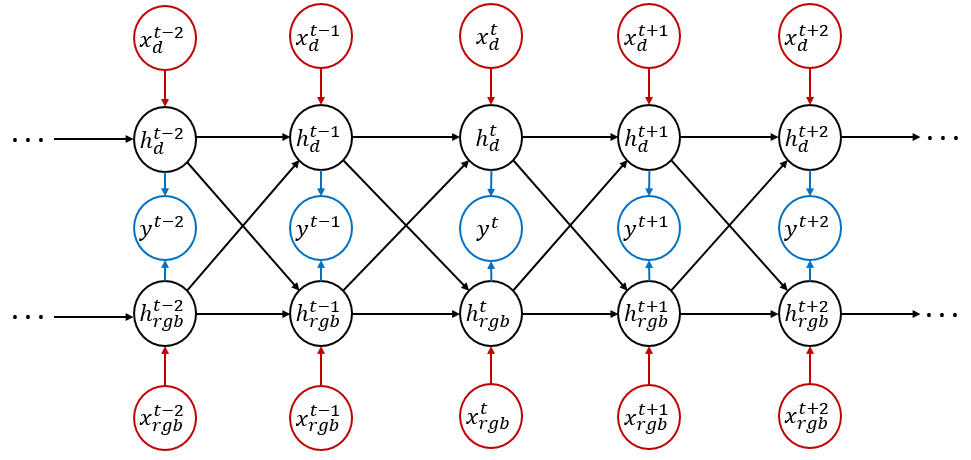}
	\end{center}
	\caption{A conceptual illustration to show our principal design in our proposed multimodal 2D-RNNs. This illustration is based on simple 1D-RNNs and intended to clarify the design structure of two crossly-connected 1D-RNNs through the introduced transfer layers. In our model, since we process 2D input signals (images) we use 2D-RNNs instead of 1D-RNN to retain the spacial dependencies within each scene image.\label{multimodal:idrnn}}
\end{figure}

We show first a conceptual illustration of our proposed model using 1D-RNNs as shown in Figure \ref{multimodal:idrnn}. Concretely, in this paper, we adapt 2D-RNN to learn hidden representations for local RGB-D image patches. Figure~\ref{2d-alex} shows one scan direction using 2D-RNN, which scans only the top-left sequence. Scanning the image in only one direction leaves some patches in the top-left sequence without being informed of the contextual information from the bottom-right patches during the forward-pass of the testing images. We approximate images using four directional 2D sequences of patches following the work \cite{ShuaiBing}. The other three directions are top-right, bottom-left and bottom-right sequences. We combine the features that are learned from these four directions to obtain the final features of the image patches. 

\subsection{Multimodal RNNs via Information Transfer Layers}
Traditional RNNs are developed to model single modality signals. In this work, we extend RNNs to represent RGB-D signal by our multimodal-RNNs, where we have a pair of single RNNs and each of them is assigned to process one modality (either RGB or Depth). Besides, we propose a transfer layer to connect the hidden planes in one RNN model and the other RNN model's hidden planes and vise versa. This transfer layers will learn to adaptively extract relevant patterns from one modality to enhance the feature representations of the other modality. If the Depth-RNN is processing the Depth patch in the sequence at location $(i,j)$, the other RNN, which is RGB-RNN, will be processing the corresponding RGB patch at the same location. The Depth-RNN is also fed with the processed hidden state values obtained from the RGB-RNN and vice versa. The RGB-D data flows concurrently in both models where their internal processing clocks are synchronized, so that at each time step we process a pair of RGB and Depth local patches simultaneously.

The architecture of our model is summarized in Figure~\ref{coupleRNNBero}. It is an end-to-end learning framework. Recurrent layers and transfer layers are automatically learned to maximize the labeling performance on the training RGB-D data. Compared to the baseline which concatenates RGB-D data and thus mixes the multimodality information (both relevant and irrelevant), our method retains modality-specific information and only shares relevant cross-modality information.

Given one 2D direction in the RNN (the first hidden plane from the quad-directional hidden planes), as shown in Figure~\ref{coupleRNNBero}, the current state of the network that is being processed at $(i,j)$ depends basically on four main previous states. Two are obtained from the network itself and the others are obtained  through the transfer layers from the previous states processed in the other modality network, in addition to the input patch features from either the RGB or the depth images at location $(i,j)$. Both networks are synchronized and process the input modalities simultaneously.

Given an RGB image $I_{c}$ where $c$ refers to `color' and is processed by RGB-RNN and a depth image $I_{d}$ where $d$ refers to `depth' and is processed by Depth-RNN, in this paper, we first extract multiple patches from the images and generate their corresponding convolutional feature vectors to form the input to our multimodal-RNNs model. We denote the corresponding convolutional feature maps as $x_{c}$ or $x_{d}$ for each patch in $I_{c}$ or $I_{d}$. Concretely, the forward propagation formulation to process one hidden plane out of the four (quad) directional hidden planes is as the following:
\begin{figure}
	\begin{center}
		\includegraphics[width=1\linewidth]{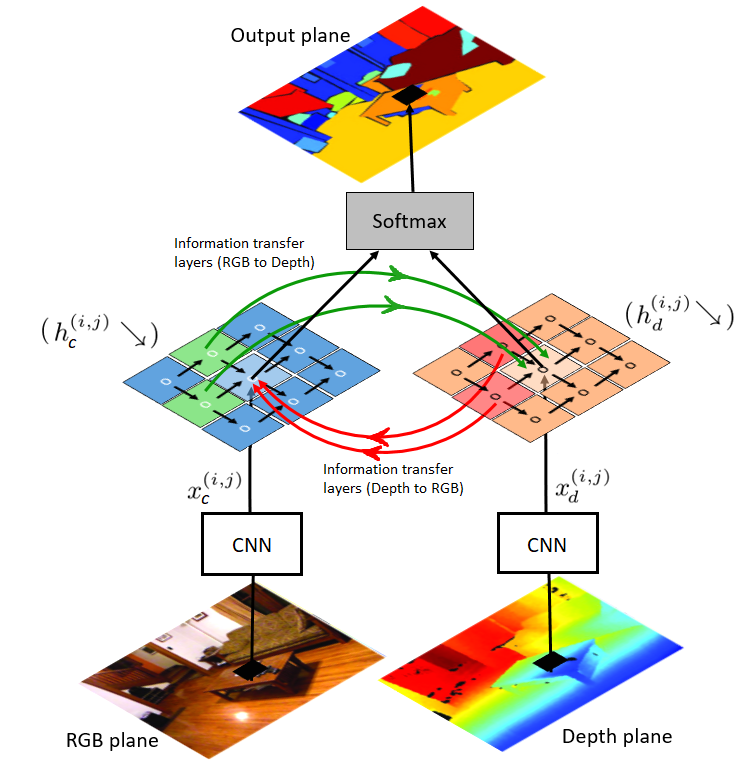}
	\end{center}
	\caption{An abstract overview of our multimodal-RNNs model which consists mainly of an RGB-RNN and a Depth-RNN. The RGB-RNN will process a patch from the RGB input plane in addition to its own previous hidden states and the selections made by the transfer layer from the previous hidden states of the Depth-RNN. The Depth-RNN will also process the corresponding patch from the Depth input plane in addition to its own previous hidden states and the selections made through the transfer layer from previous hidden states of RGB-RNN. For the simplicity in this figure, we only show one hidden plane from the quad-directional RNN hidden planes. Every hidden plane is crossly connected to its peer in the multimodal-RNNs through information transfer layers.}
	\label{coupleRNNBero}
\end{figure}

\begin{equation}
\label{one-direction}
\begin{aligned}
h_{c}^{(i,j)} &= f(U_{c} x_{c}^{(i,j)} + W_{c} h_{c}^{(i-1,j)} + W_{c} h_{c}^{(i,j-1)} \\ & + \boldsymbol{S_{c}} h_{d}^{(i-1,j)} + \boldsymbol{S_{c}} h_{d}^{(i,j-1)} + b_{c}) \\
h_{d}^{(i,j)} &= f(U_{d}x_{d}^{(i,j)} + W_{d}h_{d}^{(i-1,j)} + W_{d} h_{d}^{(i,j-1)} \\ & +
\boldsymbol{S_{d}} h_{c}^{(i-1,j)} + \boldsymbol{S_{d}} h_{c}^{(i,j-1)}+ b_{d}) \\
z_{c}^{(i,j)} &= g(V_{c} h_{c}^{(i,j)} + c_{c}) \\
z_{d}^{(i,j)} &= g(V_{d} h_{d}^{(i,j)} + c_{d})
\end{aligned}
\end{equation}
where $x_{c}^{(i,j)}$ is a feature vector of a certain patch at location $(i,j)$ in the RGB image and $x_{d}^{(i,j)}$ is a feature vector of a certain patch at location $(i,j)$ in the Depth image. $h_{c}^{(i,j)}$ and $h_{d}^{(i,j)}$ are the hidden sates inside each RGB-RNN and Depth-RNN respectively. The weight matrices $U_{c}$ and $U_{d}$ are responsible to for input-hidden mapping in RGB and Depth modalities respectively. In the other hand, here we have two types of hidden-hidden transformation matrices; within-modality hidden-hidden transformation and across-modality hidden-hidden transformation. $W_{c}$ and $W_{d}$ are within-modality hidden-hidden mapping inside the RGB-RNN and the Depth-RNN respectively. Meanwhile, $\boldsymbol{S_{c}}$ is a transformation weight matrix to transform features from the Depth to RGB hidden states (from the Depth-RNN modality hidden state to the RGB-RNN modality hidden state). $\boldsymbol{S_{d}}$ is a transformation weight matrix to transform features from the RGB to Depth hidden states (from the RGB-RNN modality hidden state to the Depth-RNN modality hidden state), i.e. these weight matrices act as the \textbf{information transfer layers that crossly connect the RGB and Depth hidden states}. Ultimately, the $V_{c}$ and $V_{d}$ are the hidden-output transformation matrices in each corresponding modality.

$\boldsymbol{S_{c}}$ and $\boldsymbol{S_{d}}$ are learnt to extract shared patterns between the modalities. Notice that the weight matrix that transforms from the top side hidden state in the RGB-RNN $h_{c}^{(i-1,j)}$ and the weight matrix that transforms from the left side hidden state $h_{c}^{(i,j-1)}$ are shared which is $W_{c}$. Similarly, $W_{d}$ in the Depth-RNN is shared (transforms from $h_{d}^{(i-1,j)}$ and from $h_{d}^{(i,j-1)}$). Also the case in the transfer layers; $\boldsymbol{S_{c}}$ and $\boldsymbol{S_{d}}$ are shared. In detail, $\boldsymbol{S_{c}}$ transforms from $h_{d}^{(i-1,j)}$ and from $h_{d}^{(i,j-1)}$. While $\boldsymbol{S_{d}}$ transforms from $h_{c}^{(i-1,j)}$ and from $h_{c}^{(i,j-1)}$.The nonlinear function $f(.)$ is ReLU $max(x,0)$ in our implementation. The function $g(.)$ is the typical softmax and $c_{c}$ and $c_{d}$ are biases.

Since we adopt the quad-directions, the forward pass is similar to Equation \ref{one-direction} and in addition to the remaining directions beside the top-left sequence, i.e. top-right, bottom-left and bottom-right. To facilitate readability of the equation and to easily distinguish between all quad/four directions; we will use arrow notations to represent different directions. $\searrow$ refers to the top-left processing sequence, $\swarrow$ indicates the top-right sequence, $\nearrow$ is the bottom-left and finally $\nwarrow$ is the bottom-right sequence. Now, the full model forward propagation pass in the RGB-RNN becomes:
\begin{equation}
	\small
	\label{coupled-forwaredPass-master}
	\begin{aligned}
		\begin{split}
			h_{c}^{(i,j)}\searrow &= f(U^{\searrow}_{c}x_{c}^{(i,j)} + W^{\searrow}_{c} h_{c}^{(i-1,j)\searrow} + W^{\searrow}_{c} h_{c}^{(i,j-1)}\searrow \\ & + \boldsymbol{S^{\searrow}_{c}} h_{d}^{(i-1,j)}\searrow + \boldsymbol{S^{\searrow}_{c}} h_{d}^{(i,j-1)}\searrow + b^{\searrow}_{c}) \\
			h_{c}^{(i,j)}\swarrow &= f(U^{\swarrow}_{c}x_{c}^{(i,j)} + W^{\swarrow}_{c} h_{c}^{(i-1,j)\swarrow} + W^{\swarrow}_{c} h_{c}^{(i,j+1)}\swarrow \\ & + \boldsymbol{S^{\swarrow}_{c}} h_{d}^{(i-1,j)}\swarrow + \boldsymbol{S^{\swarrow}_{c}} h_{d}^{(i,j+1)}\swarrow + b^{\swarrow}_{c}) \\
			h_{c}^{(i,j)}\nearrow &= f(U^{\nearrow}_{c}x_{c}^{(i,j)} + W^{\nearrow}_{c} h_{c}^{(i+1,j)\nearrow} + W^{\nearrow}_{c} h_{c}^{(i,j-1)}\nearrow \\ & + \boldsymbol{S^{\nearrow}_{c}} h_{d}^{(i+1,j)}\nearrow + \boldsymbol{S^{\nearrow}_{c}} h_{d}^{(i,j-1)}\nearrow + b^{\nearrow}_{c}) \\
			h_{c}^{(i,j)}\nwarrow &= f(U^{\nwarrow}_{c}x_{c}^{(i,j)} + W^{\nwarrow}_{c} h_{c}^{(i+1,j)\nwarrow} + W^{\nwarrow}_{c} h_{c}^{(i,j+1)}\nwarrow \\ & + \boldsymbol{S^{\nwarrow}_{c}} h_{d}^{(i+1,j)}\nwarrow + \boldsymbol{S^{\nwarrow}_{c}} h_{d}^{(i,j+1)}\nwarrow + b^{\nwarrow}_{c}) \\
			z_{c}^{(i,j)} &= g(V^{\searrow}_{c} h_{c}^{(i,j)} \searrow + V^{\swarrow}_{c} h_{c}^{(i,j)} \swarrow + V^{\nearrow}_{c} h_{c}^{(i,j)} \nearrow \\ &  + V^{\nwarrow}_{c} h_{c}^{(i,j)} \nwarrow + c_{c})
		\end{split}
	\end{aligned}
\end{equation}

As mentioned, we use four quad-directional 2D-RNN to approximate each image. The arrows $\searrow, \swarrow, \nearrow, \nwarrow$ indicate the quad-directions (top-left, top-right, bottom-left and bottom-right), and the $h_{c}^{(i,j)}\searrow$ , $h_{c}^{(i,j)}\swarrow$ , $h_{c}^{(i,j)}\nearrow$ and $h_{c}^{(i,j)}\nwarrow$ are the corresponding four hidden planes. Each hidden plane has its own weight matrices besides the introduced transfer layers $\boldsymbol{S_{c}}$, e.g. $U^{\searrow}_{c}$ (input-hidden mapping), $W^{\searrow}_{c}$ (hidden-hidden mapping), $S^{\searrow}_{c}$ (the transfer layer from Depth-RNN hidden plane to RGB-RNN hidden plane), $V^{\searrow}_{c}$ (hidden-output mapping) and the bias term which is denoted by $b^{\searrow}_{c}$. Note that the connection between the two hidden planes $h_{d}\searrow$ and the corresponding $h_{c}\searrow$ is weighted by the corresponding weight matrix $S^{\searrow}_{c}$, which is leaned to extract shared patterns between the modalities.

\textbf{Simultaneously}, the forward pass in the Depth-RNN model is as follows:
\begin{equation}
	\small
	\label{coupled-forwaredPass-slave}
	\begin{aligned}	
		\begin{split}
			h_{d}^{(i,j)}\searrow &= f(U^{\searrow}_{d}x_{d}^{(i,j)} + W^{\searrow}_{d}h_{d}^{(i-1,j)}\searrow + W^{\searrow}_{d} h_{d}^{(i,j-1)}\searrow \\ & +  \boldsymbol{S^{\searrow}_{d}} h_{c}^{(i-1,j)}\searrow + \boldsymbol{S^{\searrow}_{d}} h_{c}^{(i,j-1)} \searrow + b^{\searrow}_{d}) \\
			h_{d}^{(i,j)}\swarrow &= f(U^{\swarrow}_{d}x_{d}^{(i,j)} + W^{\swarrow}_{d}h_{d}^{(i-1,j)}\swarrow + W^{\swarrow}_{d} h_{d}^{(i,j+1)}\swarrow \\ & + \boldsymbol{S^{\swarrow}_{d}} h_{c}^{(i-1,j)}\swarrow + \boldsymbol{S^{\swarrow}_{d}} h_{c}^{(i,j+1)} \swarrow + b^{\swarrow}_{d}) \\
			h_{d}^{(i,j)}\nearrow &= f(U^{\nearrow}_{d}x_{d}^{(i,j)} + W^{\nearrow}_{d}h_{d}^{(i+1,j)}\nearrow + W^{\nearrow}_{d} h_{d}^{(i,j-1)}\nearrow \\ & + \boldsymbol{S^{\nearrow}_{d}} h_{c}^{(i+1,j)}\nearrow + \boldsymbol{S^{\nearrow}_{d}} h_{c}^{(i,j-1)} \nearrow + b^{\nearrow}_{d}) \\
			h_{d}^{(i,j)}\nwarrow &= f(U^{\nwarrow}_{d}x_{d}^{(i,j)} + W^{\nwarrow}_{d}h_{d}^{(i+1,j)}\nwarrow + W^{\nwarrow}_{d} h_{d}^{(i,j+1)}\nwarrow \\ & + \boldsymbol{S^{\nwarrow}_{d}} h_{c}^{(i+1,j)}\nwarrow + \boldsymbol{S^{\nwarrow}_{d}} h_{c}^{(i,j+1)} \nwarrow + b^{\nwarrow}_{d}) \\
			z_{d}^{(i,j)} &= g(V^{\searrow}_{d} h_{d}^{(i,j)} \searrow + V^{\swarrow}_{d} h_{d}^{(i,j)} \swarrow + V^{\nearrow}_{d} h_{d}^{(i,j)} \nearrow \\ &  + V^{\nwarrow}_{d} h_{d}^{(i,j)} \nwarrow + c_{d})
		\end{split}
	\end{aligned}
\end{equation}
where $x^{(i,j)}_{d}$ is the feature vector of a certain patch at location $(i,j)$ in the depth image. The $h_{d}^{(i,j)}\searrow$, $h_{d}^{(i,j)}\nearrow$, $h_{d}^{(i,j)}\swarrow$ and $h_{d}^{(i,j)}\nwarrow$ are the quad hidden planes in the Depth-RNN. Each hidden plane accompanies its own weight matrices $U^{\searrow}_{d}$ (input-hidden mapping), $W^{\searrow}_{d}$ (hidden-hidden mapping), $S^{\searrow}_{d}$ (transfer layer from RNN-RGB hidden plane to Depth-RNN hidden plane), $V^{\searrow}_{d}$ (hidden-output mapping) and its own bias term $b^{\searrow}_{d}$. The remaining terms in the quad planes are similar to the case of the first hidden plane. The function $f(.)$ is a nonlinear ReLU unit and the function $g(.)$ is the typical softmax and $c_{d}$ is a bias term.

We can notice that the cross-connections through the transfer layers are applied in each processing direction (four sequences). By this method, the processing of a specific patch will rely on both previous hidden neighbors from its own modality in addition to the other modalities, thus learns more contextually-aware hidden representations of the RGB-D image patches from both modalities.

Our labeling task is a typical supervised classification problem. We aggregate the cross entropy losses from both modalities and calculate the loss for every patch. The error signal of an image is averaged across all the valid patches (those that are semantically labeled), which is mathematically formulated as the following:
\begin{equation}
	\small
	\label{coupled-lossCrossEnrtopy}
	L = -\frac{1}{N}\sum_{n=1}^N\sum_{b=1}^{B} (log ({z}^n_c (b) \ ) \delta({y}^{n}=b) + log ({z}^n_{d} (b) \ ) \delta({y}^{n}=b))
\end{equation}
where $\delta(\cdot)$ is the indicator function, $N$ is the total number of patches and $B$ is the number of semantic classes, $y^n$ is the ground truth label for the RGB-D patch representation, which is the same as that of the center pixel, ${z}^n_{c}$ and  ${z}^n_{d}$ are the class likelihood generated from both the RGB-RNN and the Depth-RNN for the patch representations $\mathbf{x}^n_{c}$ and  $\mathbf{x}^n_{d}$ respectively, where they are $B$-dimensional vectors. Note that we ignore the contribution of unlabeled (invalid) patches in the loss calculation.

\textbf{Optimization of the multimodal-RNNs model:}
To learn the multimodal RNNs parameters in Equations \ref{coupled-forwaredPass-master} and \ref{coupled-forwaredPass-slave}, we optimize the objective function in Equation~\ref{coupled-lossCrossEnrtopy} with a stochastic gradient-based method. Both RGB-RNN and Depth-RNN are optimized simultaneously using Back Propagation Through Time (BPTT) \cite{1D-RNN-Elman}. We unfold both networks in time and calculate their gradients which are back-propagated at each time step throughout both networks. This is similar to the typical multilayer feed-forward neural network propagation, but with the difference that the weights are shared across time steps defined by the architecture of the recurrent network. The whole model is differentiable and trained end-to-end. The propagation of the gradients through the RGB-RNN and Depth-RNN is simultaneous.

In more detail, we provide the first plane (top-left sequence) backward pass in the RGB-RNN. The backward pass derivations of the remaining sequences (the remaining three planes) can be easily inferred by following similar derivation strategy as we will explain next in the backward pass formulations for the first plane, but with considering the change of the sequence directions (top-right, bottom-left and bottom-right sequences). Similarity, the total derivations for the Depth-RNN are straightforward and exactly the same as for the RGB-RNN but with swapping the notations of $c$ and $d$ and vise versa.

In the RGB-RNN top-left sequence, we generate the gradients of the loss function in Equation \ref{coupled-lossCrossEnrtopy} by deriving it with respect to the model internal parameters, i.e. top-left sequence weight matrices $U^{\searrow}_{c}$, $W^{\searrow}_{c}$, $\boldsymbol{S^{\searrow}_{c}}$,   $V^{\searrow}_{c}$, $b^{\searrow}_{c}$ and $c_{c}$. 

Notice that since the weight matrix $W^{\searrow}_{c}$ is shared between $h_{c}^{(i-1,j)\searrow}$ and $h_{c}^{(i,j-1)}\searrow$ and the transfer layer weight matrix $\boldsymbol{S^{\searrow}_{c}}$ is shared between $h_{d}^{(i-1,j)}\searrow$ and $h_{d}^{(i,j-1)} \searrow$, hence we will rewrite the forward pass as the following to further facilitate the understanding of the weight sharing concept and the backward pass:

\begin{equation}
	\label{coupled-forwaredPass-master-rgb-packed}
	\begin{aligned}
		\begin{split}
		    \tilde{h_{c}}\searrow &= h_{c}^{(i-1,j)\searrow} + h_{c}^{(i,j-1)}\searrow \\
		    \tilde{h_{d}}\searrow &= h_{d}^{(i-1,j)\searrow} + h_{d}^{(i,j-1)}\searrow \\
			h_{c}^{(i,j)}\searrow &= f(U^{\searrow}_{c}x_{c}^{(i,j)} + W^{\searrow}_{c} \tilde{h_{c}}\searrow + \boldsymbol{S^{\searrow}_{c}} \tilde{h_{d}}\searrow + b^{\searrow}_{c}) \\
			z_{c}^{(i,j)} &= g(V^{\searrow}_{c} h_{c}^{(i,j)} \searrow + c_{c})
		\end{split}
	\end{aligned}
\end{equation}

Notice that the other remaining terms that present originally in Equation \ref{coupled-forwaredPass-master} (i.e. $V^{\swarrow}_{c} h_{c}^{(i,j)}\swarrow$, $V^{\nearrow}_{c} h_{c}^{(i,j)}\nearrow$, and $V^{\nwarrow}_{c} h_{c}^{(i,j)}\nwarrow)$ do not engage any of the internal parameters of the top-left sequence, hence we omit them.

\begin{figure}
	\centering
	\begin{center}
		\includegraphics[width=0.45\textwidth]{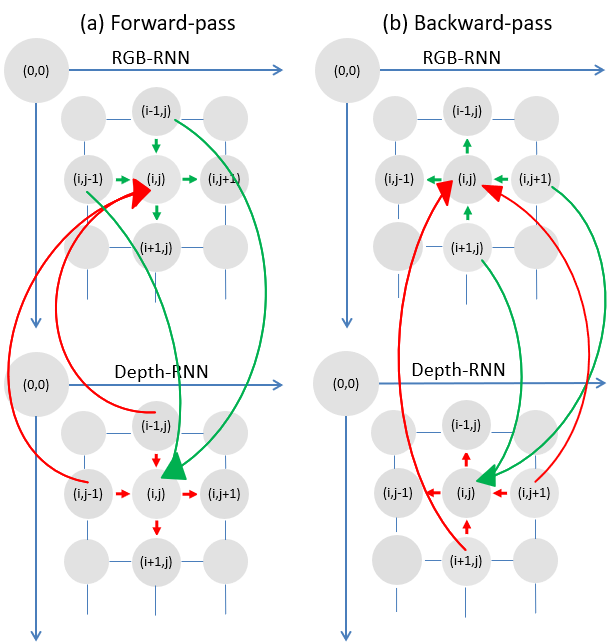}
	\end{center}
	\caption {An illustration of the forward and backward passes in the top-left plane sequence in both RGB-RNN and Depth-RNN. (Best viewed in color)}
	\label{FBP}
\end{figure}

Before we formulate the backward pass, Figure \ref{FBP} show an illustration of the forward and backward passes in the top-left plane sequence. Notice that the derivatives which are computed in the backward pass at each hidden state at some specific location $(i,j)$ are processed in the \textbf{reverse order} of forward propagation sequence.

\textbf{For better readability, we ignore the south-east arrow sign $\searrow$ in the following derivations for the top-left sequence}. 

Notice that now there are \textbf{two} types of error signals: \textbf{direct} one which is reachable directly through the loss ($\dfrac{\partial z_{c}^{(i,j)}}{\partial h_{c}^{(i,j)}}$) and \textbf{indirect} ones, i.e. the error signals that are coming from neighboring future states at $(i+1,j)$ and $(i,j+1)$ locations; ($\dfrac{\partial z_{c}^{(i+1,j)}}{\partial h_{c}^{(i+1,j)}}\dfrac{\partial h_{c}^{(i+1,j)}}{\partial h_{c}^{(i,j)}}$ + $\dfrac{\partial z_{c}^{(i,j+1)}}{\partial h_{c}^{(i,j+1)}}\dfrac{\partial h_{c}^{(i,j+1)}}{\partial h_{c}^{(i,j)}}$).

Concretely, and given that the derivative of loss function $L$ with respect to the Softmax output function $g$ is $g'(.) = \dfrac{\partial L}{\partial z_{c}^{(i,j)}(.)} \dfrac{\partial z_{c}^{(i,j)}(.)}{\partial g}$, similarity 
$f'(.) = \dfrac{\partial h_{c}^{(i,j)}}{\partial f}$, hence at location $(i,j)$ the backward pass of the RGB-RNN (derivations w.r.t. all internal parameters) is formulated as the following:
\begin{equation}
	\label{coupled-backwardPass-rgb}
	\begin{aligned}
	\Delta V_{c} &= g'(z^{(i,j)}_{c}) (h_{c}^{(i,j)})^T \\
 	dh^{(i,j)}_{c} &= V_{c}^T g'(z^{(i,j)}_{c}) +  W_{c}^T dh^{(i+1,j)}_{c} \circ f'(h_{c}^{(i+1,j)}) \\& + W_{c}^T dh^{(i,j+1)}_{c} \circ f'(h_{c}^{(i,j+1)}) \\
	\Delta U_{c} &= dh^{(i,j)}_{c} \circ f'(h_{c}^{(i,j)}) (x_{c}^{(i,j)})^{T} \\
	\Delta W_{c} &= dh^{(i+1,j)}_{c} \circ f'(h_{c}^{(i+1,j)}) (h_{c}^{(i,j)})^{T} \\& + dh^{(i,j+1)}_{c} \circ f'(h_{c}^{(i,j+1)}) (h_{c}^{(i,j)})^{T} \\
	\Delta \boldsymbol{S_{c}} &= dh^{(i+1,j)}_{c} \circ f'(h_{c}^{(i+1,j)}) (h_{d}^{(i,j)})^{T} \\& + dh^{(i,j+1)}_{c} \circ f'(h_{c}^{(i,j+1)}) (h_{d}^{(i,j)})^{T} \\
	\Delta b_{c} &= dh^{(i,j)}_{c} \circ f'(h_{c}^{(i,j)})\\
	\Delta c_{c} &= g'(z^{(i,j)}_{c})\\
	\end{aligned}
\end{equation}
The term $dh^{(i,j)}_{c}$ is defined to allow the model to propagate local information internally. The $\circ$ sign is the Hadamard product (element-wise).  We can notice through the term $dh^{(i,j)}_{c}$ that there are two sources of gradients, one is generated directly from the current hidden state and the other generated indirectly from the bottom and right locations (corresponding to the three terms: direct error: $V_{c}^T g'(z^{(i,j)}_{c})$, indirect error from neighbor bottom location: $W_{c}^T dh^{(i+1,j)}_{c} \circ f'(h_{c}^{(i+1,j)})$ and indirect from right neighbor location: $W_{c}^T dh^{(i,j+1)}_{c} \circ f'(h_{c}^{(i,j+1)})$). Notice that these two types of error signals (direct and indirect) are due to the weight sharing effect. It is also important to mention that in our derivations we show the error signals coming from the direct bottom and right neighbors, however, in real implementations the equations are recurrently applied to allow propagation of the errors recursively (recurrently) from all potential future neighbors until we reach out the current referenced patch at specific location.

Similarly, the backward passes for the remaining three planes in the RGB-RNN are straightforward to be derived following Equation \ref{coupled-backwardPass-rgb} but with paying attention to the directions of the processing sequence. Additionally, the backward pass in the Depth-RNN follows exactly similar derivations but with swapping the notations of $c$ with $d$ and vise versa. Notice that the derivation w.r.t $c_{c}$ is the same in the other remaining three planes/sequences.

To recap, we adopt BPTT to train both basic quad-directional 2D-RNN (RGB-RNN and Depth-RNN). We use the cross entropy loss in our implementation and the errors are calculated by chain rule as we described previously in detail.

\section{Experiments and Results}
\subsection{Datasets}We evaluate our model on the benchmark dataset NYU versions 1 and 2 \cite{Silberman:indoorSegmentationv1,Silberman:indoorSegmentationv2}. The NYU V1 dataset contains 2284 RGB-D indoor scene images labeled with around 13 categories. The NYU V2 is also comprised of video sequences from indoor scenes as recorded by both the RGB and Depth cameras from the Microsoft Kinect. It contains 1449 densely labeled pairs of aligned RGB and depth images. We follow the settings in \cite{Silberman:indoorSegmentationv2}, where the first task is to predict the pixel label out of four semantic classes: Ground, Furniture, Props and Structure. The second task is to predict the label out of 14 categories. The accuracy is calculated in terms of total pixel-wise, average class-wise and Intersection Over Union (average IOU) among the semantic classes for comparison.

\subsection{CNNs and RNNs Training}We train our convolutional neural networks (CNNs) on both NYU version 1 and 2 images. We follow the network structure proposed by \cite{ShuaiBing} without considering any spatial information channels (RGB locations). We also didn't perform any hybrid sampling as they mention in their implementation details. The CNN model consists mainly of three convolutional layers (where in between there are max-pooling and ReLU layers) and two last fully-connected layers followed by an $B$-way softmax loss layer. In more details, conv1($8\times8\times3$ or $4\times32$), max pool($2\times2$), conv2($6\times6\times32\times64$), max pool($2\times2$), conv3($5\times5\times64\times64$), max pool($2\times2$), FC1($1024\times64$) and FC2($64\times B$). The CNN models are trained on $65 \times 65$ patches associated with their centering pixel labels. Each image contains $4661$ patches. Each CNN model produces a 64-dimensional vector per patch for each modality. We use these CNN features as the input of our RNN models. 

In CNN training, we start by a learning rate of $0.001$ and decrease it every $5$ epochs by $10$ (divided by $10$). The momentum is initialized as $0.9$ and remains the same throughout the training. We perform the typical normalization as a preprocessing step of images, we subtract the mean image and divide by the standard deviation. The CNN models nearly converge after 50 epochs (around 6 hours on NVIDIA TK40 GPU), however, we iterate the models in most cases until 100 epochs. 

We train our multimodal RNNs using BPTT (back propagation through time). We adapt stochastic gradient descent (SGD) throughout our training. We initialize the learning rate as $0.00001$ and decrease it by $0.01$ after each epoch. The momentum is set to $0.9$. The internal parameter dimension of the networks is set to $64$. We apply gradient clipping \cite{advancesOptimization,difficultyOptimization} and set the threshold value to $2 * 10^{3}$, however, even if the ReLU can potentially cause gradient explosion, it plays a critical role in RNN to mitigate the gradient vanishing problem. Thus, we mainly use ReLU in all of our RNN models. The other RNN model parameters are initialized either by randomly or zeros. Notice that we didn't consider any pixels that have zero labels in our training, we held them out. Our model converges much faster compared to the other competitive baselines. It converged at almost the similar speed of a single RNN model, with the benefit of processing multiple input modalities simultaneously.
\subsection{Baselines}
To show the effectiveness of our proposed model, we develop the following baselines for comparison alongside our proposed model:
\begin{itemize}
	\item \textbf{CNN-RGB:} in this baseline, we train CNN based on RGB images (input is three RGB channels) for label prediction.
	
	\item \textbf{CNN-Depth:} we train CNN as in `CNN-RGB' but using Depth images only (input is one Depth channel).
	
	\item \textbf{CNN-RGBD:} we train CNN as in `CNN-RGB' with extra Depth images (input is four RGB-D channels) similar to the work presented in \cite{indoorSemanticYannLecun}.
	
	\item \textbf{RNN-RGB:} in this baseline, we follow the structure of the quad-directional 2D-RNN proposed by \cite{ShuaiBing}. We use the model in `CNN-RGB' to extract RGB features. We only input the RGB features to train the RNN for label prediction.
	
	\item \textbf{RNN-Depth: } we extract Depth features using the trained model in `CNN-Depth' and use these features to train quad 2D-RNN as in `RNN-RGB'. We only input the Depth features to train the RNN for label prediction.
	
	\item \textbf{RNN-RGBD: } we use the trained model in `CNN-RGBD' for feature extraction. We use these RGB-D features to train quad 2D-RNN similar to `RNN-RGB' baseline to perform label prediction.
	
	\item \textbf{RNN-Classifiers-Combined: } or as we call it `post-fusion', here we train RGB-RNN and Depth-RNN for label prediction and combine their classification scores on the classifier level. We use the trained models in `CNN-RGB' and `CNN-Depth' for feature extraction. 
	
	\item \textbf{RNN-Features-Combined: } or `pre-fusion', here we use both trained models in `CNN-RGB' and `CNN-Depth' for feature extraction. We concatenate both RGB and Depth features (to form higher dimensional feature vectors) and train one quad 2D-RNN similar to `RNN-RGB' baseline to perform label prediction.
	
	\item \textbf{RNN-Hiddens-Combined: } or `middle-fusion', we fuse the hidden representations of both RNNs just before classification and train them jointly.
	
	\item \textbf{Multimodal-RNNs-Ours} in this sitting we implement our proposed  multimodal RNNs structure. Here, we have two internal RNN models one is responsible for processing the RGB features and the other is responsible for processing the Depth features. Both models are optimized simultaneously using BPTT where we combine their classification scores to finally obtain the label map per RGB-D image.
	
	\item \textbf{Multimodal-RNNs-Ours-Multiscale: } similar to our `Multimodal-RNNs-Ours', but in this setting we applied our proposed structure on multiscale convolutional features as proposed by \cite{sceneParsingPurityTrees}. We follow similar training sittings to train multiscale CNN models on different image sizes, then we use them to extract multiscale features. We concatenate these features together to form our final input patch representation to our multimodal-RNNs.
	
	We also compare our results with other state-of-the-art methods.
\end{itemize}

\subsection{Results}
Given the input RGB images and their corresponding Depth images, we divide each image into non-overlapping patches of size $65\times65$. The extracted local CNN features are used as the input of our multimodal RNNs. The RGB-RNN and the Depth-RNN models process RGB and Depth local patch features respectively.

\begin{figure*}
	\centering
	\begin{center}
		\includegraphics[width=0.8\textwidth]{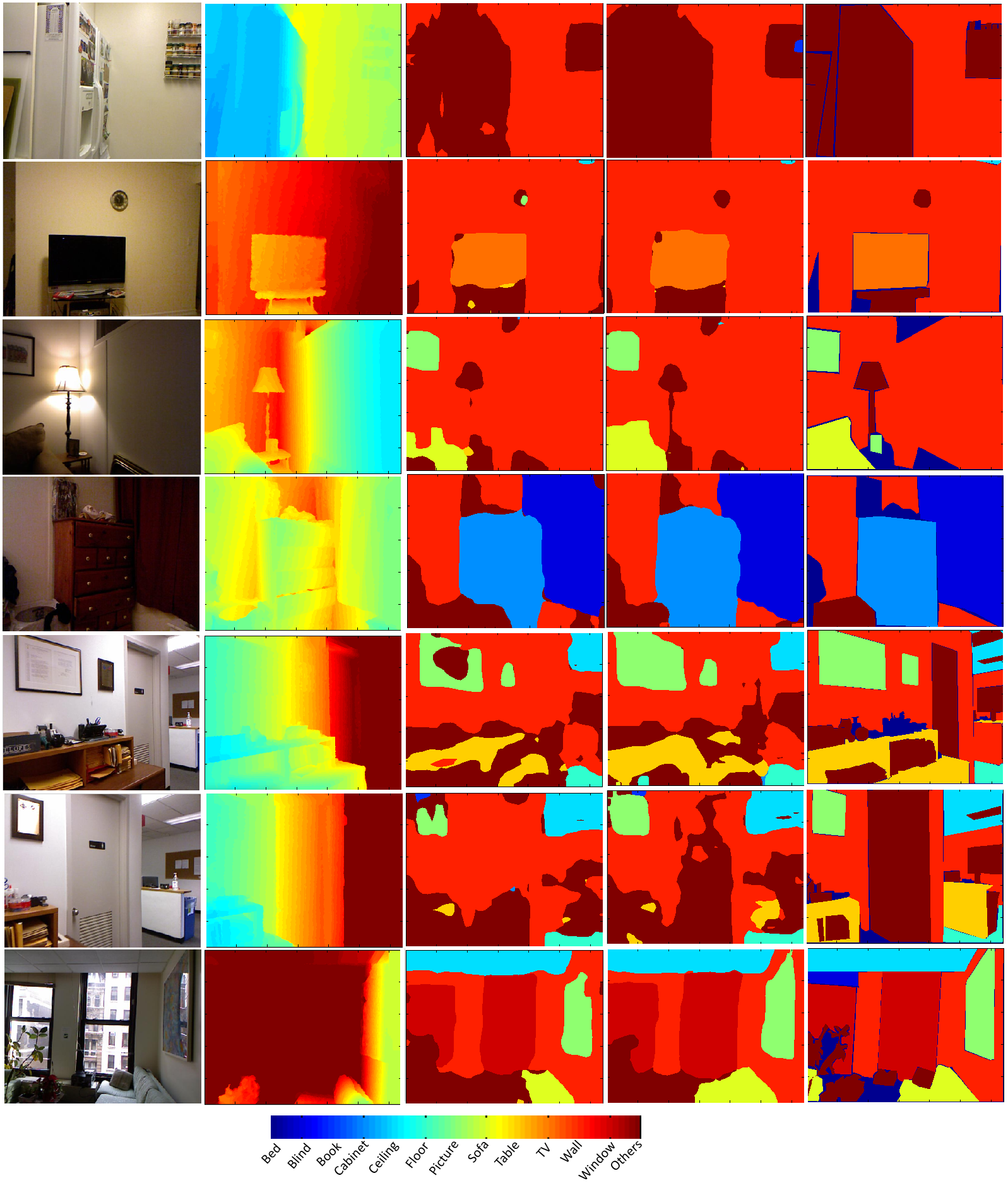}
	\end{center}
	\caption {$7$ example results from NYU-V1 (13 semantic categories). Columns 1st and 2nd: RGB color images and their corresponding Depth images. Column 3rd: the prediction results of our most competitive baseline `RNN-Features-Combined' on the NYU V2 four semantic categories. Column 4th: our multimodal-RNNs prediction results. Column 5th:ground truth. Our model is able to correctly classify many miss-classified pixels compared with our baseline. (Best viewed in color)}
	\label{missClassification2}
\end{figure*}

\textbf{Results on NYU V1 - 13 categories: }
In this dataset the task is to predict pixel labels out of 12 categories plus an unknown category. Table~\ref{TableV1} shows the results of our baselines alongside the multimodal RNNs model and other state-of-the-art methods. RNN models outperform CNN baselines by a large margin. And our multimodal RNNs further improve the accuracy over the RNN baselines. Our model achieves comparable results with the other state-of-the-art methods. The accuracy comparison with all baselines shows the effectiveness of our method with the proposed transfer layers. Figure \ref{missClassification2} show some qualitative results generated by our method and the most competitive baseline method `RNN-Features-Combined'. Our multi-modal-RNNs model can correctly classify many mis-classifications results compared to the baseline in most cases.

\begin{table}
	\begin{center}
		\begin{tabular}{|l|c|c|c|}
			\hline
			Algorithm & Pixel Acc & Class Acc & IOU\\
			\hline\hline
			CNN-RGB & 69.21\% & 56.23\% & 45.17\%\\
			CNN-Depth & 58.56\% & 32.82\% & 22.21\%\\
			CNN-RGBD & 69.17\% & 57.83\% & 45.24\% \\
			\hline
			RNN-RGB & 71.38\% & 64.43\% & 52.80\%\\
			RNN-Depth & 60.67\% & 46.88\% & 28.24\%\\
			RNN-RGBD & 71.09\% & 67.01\% & 57.87\%\\
			RNN-Features-Combined & 72.94\% & 69.06\% & 60.54\%\\
			RNN-Hiddens-Combined  & 71.50\% & 67.00\% & 55.59\%\\
			RNN-Classifiers-Combined & 71.28\% & 66.37\% & 54.79\% \\
			\hline
			Multimodal-RNNs-Ours & 74.68\% & 72.54\% & 62.53\% \\
			Multimodal-RNNs-Ours-Multiscale & \textbf{78.89}\% & \textbf{75.73}\% & \textbf{65.70}\% \\
			\hline
			Wang et al.\cite{WangAnranUnsupervised} & - & 61.71\%  & -\\
			gradient KDES\cite{rgbdSceneFeatures} & - &  51.84\%  & -\\
			color KDES\cite{rgbdSceneFeatures} & - & 53.27\%  & -\\
			spin/surface normal KDES\cite{rgbdSceneFeatures}  & - & 40.28\%  & -\\
			depth gradient KDES\cite{rgbdSceneFeatures} &  - & 53.56\% & - \\
			Silberman et al\cite{Silberman:indoorSegmentationv1} &  - & 53.00\%  & -\\
			Pei et al.\cite{peiUnsupervisedLearning} &  - & 50.50\%  & -\\
			Ren et al.\cite{rgbdSceneFeatures}  & - & 71.40\%  & -\\
			Eigen et al. \cite{predictinDpethCVPR2015} & 75.40\% & 66.90\% & - \\
			\hline
		\end{tabular}
	\end{center}
	\caption{Results on \textbf{NYU V1} dataset \cite{Silberman:indoorSegmentationv1}. The performance of all baselines and other state-of-the-art methods is measured on total pixel-wise, average class-wise accuracy and average IOU (intersection over union) among \textbf{$13$} semantic classes. The higher the better.}
	\label{TableV1}
\end{table}

\textbf{Results on NYU V2 - 4 categories: }
For example, Eigen et al. \cite{predictinDpethCVPR2015} achieve higher results than our model on this task, while we outperform their model on NYU V1 task. Their multi-scale CNN structure is also well designed to address the problem. However, their network appears to be much more complex than ours as they have higher number of computational operations (way more convolutions).  But we still believe our work can be potentially orthogonal and complementary to their method if it is trained end-to-end with their method.

\begin{figure*}
	\centering
	\begin{center}
		\includegraphics[width=0.8\textwidth]{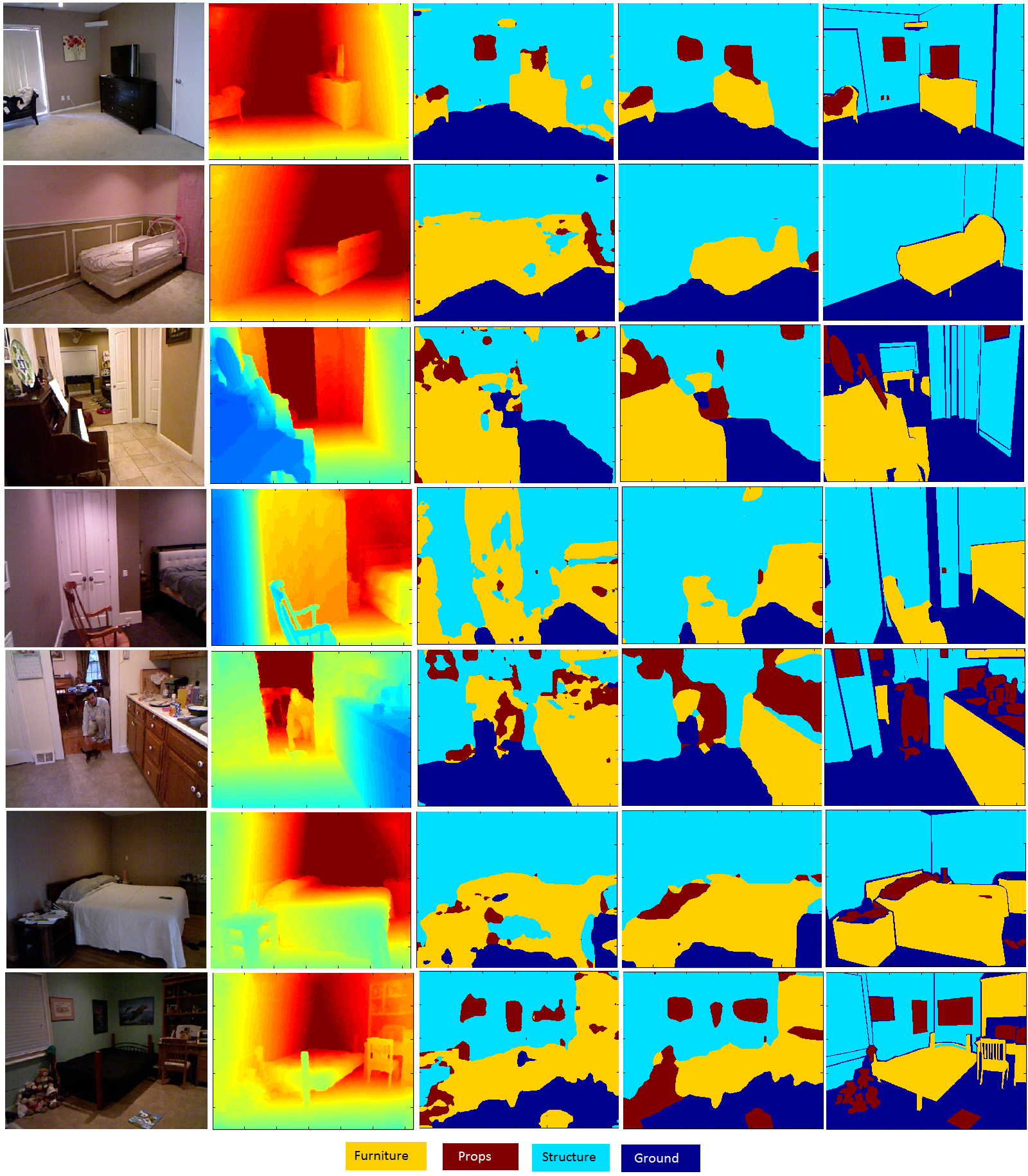}
	\end{center}
	\caption {$7$ example results. Columns 1st and 2nd: RGB color images and their corresponding Depth images. Column 3rd: the prediction results of our most competitive baseline `RNN-Features-Combined' on the NYU V2 four semantic categories. Column 4th: our multimodal-RNNs prediction results. Column 5th:ground truth. Our model is able to correctly classify many miss-classified pixels compared with our baseline. (Best viewed in color)}
	\label{missClassification1}
\end{figure*}

Likewise, Couprie et al.\cite{Silberman:indoorSegmentationv2}, use many types of features, including SIFT features, histograms of surface normals, 2D and 3D bounding box dimensions, color histograms, relative depth and their support features. In our model, we only use the transfer layers alongside the basic quad 2D-RNN structures, and can achieve much better performance. Figure \ref{missClassification1} also show some qualitative results generated by our method and the most competitive baseline method `RNN-Features-Combined'. Our multi-modal-RNNs model can correctly classify many mis-classifications results compared to the baseline in most cases.
\begin{table}
	\begin{center}
		\begin{tabular}{|l|c|c|c|}
			\hline
			Algorithm & Pixel Acc & Class Acc & IOU\\
			\hline\hline
			CNN-RGB & 65.55\% & 62.07\% & 45.42\%\\
			CNN-Depth & 69.61\% & 65.96\% & 49.13\%\\
			CNN-RGBD & 71.60\% & 69.89\% & 54.19\%\\
			\hline
			RNN-RGB & 68.14\% &  65.96\% & 51.05\%\\
			RNN-Depth & 70.95\% & 67.76\% & 52.21\%\\
			RNN-RGBD & 74.18\% & 69.99\% & 56.63\%\\
			RNN-Features-Combined & 74.36\% & 72.80\% & 60.08\%\\
			RNN-Hiddens-Combined  & 73.70\% & 71.10\% & 56.00\%\\
			RNN-Classifiers-Combined & 73.39\% & 70.64\% & 55.34\%\\
			\hline
			Multimodal-RNNs-Ours & 75.74\% & 75.01\%  & 62.10\%\\
			Multimodal-RNNs-Ours-Multiscale & 78.60\% & 76.69\% & \textbf{65.09}\% \\
			\hline
			Wang et al.\cite{WangAnranUnsupervised} & - & 65.30\% & -\\
			Couprie et al. \cite{indoorSemanticYannLecun}   & 64.50\% & 63.50\% & -\\
			Stuckler et al.\cite{stucklerMapping} & 70.90\% & 67.00\% & -\\
			Khan et al. \cite{salmanKhanGeometry} & 69.20\% & 65.60\% & -\\
			Mueller et al.\cite{mullerDepthCRF} & 72.30\% & 71.90\% & -\\
			Gupta et al.\cite{gupta2013cvpr} & 78.00\% & - & 64.00\%\\
			Cadena and Kosecka \cite{cadenaKosckaRGBNYU} & - & 64.10\% & -\\
			Eigen et al. \cite{predictinDpethCVPR2015} & \textbf{83.20\%} & \textbf{82.00\%} & - \\
			
			\hline
		\end{tabular}
	\end{center}
	\caption{Results on \textbf{NYU V2} dataset \cite{Silberman:indoorSegmentationv2}. The performance of all baseline and other state-of-the-art methods is measured on total pixel-wise, average class-wise accuracy and average IOU (intersection over union) among \textbf{$4$} semantic classes: Ground, Furniture, Props and Structure. The higher the better.}
	\label{TableV2-4}
\end{table}

\textbf{Results on NYU V2 - 14 categories:}
On the NYU V2, we evaluate our model to label image pixels with one of 13 categories plus an unknown category. We show the comparison in this task with various baselines and state-of-the-art methods as shown in Table~\ref{TableV2-14}. This is the most competitive benchmark presented on this dataset. Notice that the work of Cadena and Kosecka \cite{cadenaKosckaRGBNYU} used many RGB image and 3D features and formulate the problem in the CRF framework. Meanwhile the work of Wang et al. \cite{WangAnranUnsupervised} adapted an existing unsupervised feature learning technique to directly learn features. They stack their basic learning structure to learn hierarchical features. They combined their higher-level features with low-level features and train linear SVM classifiers to perform labeling. Compared to these methods, our model is much simpler and achieve better performance.

\begin{figure*}
	\centering
	\begin{center}
		\includegraphics[width=0.85\textwidth]{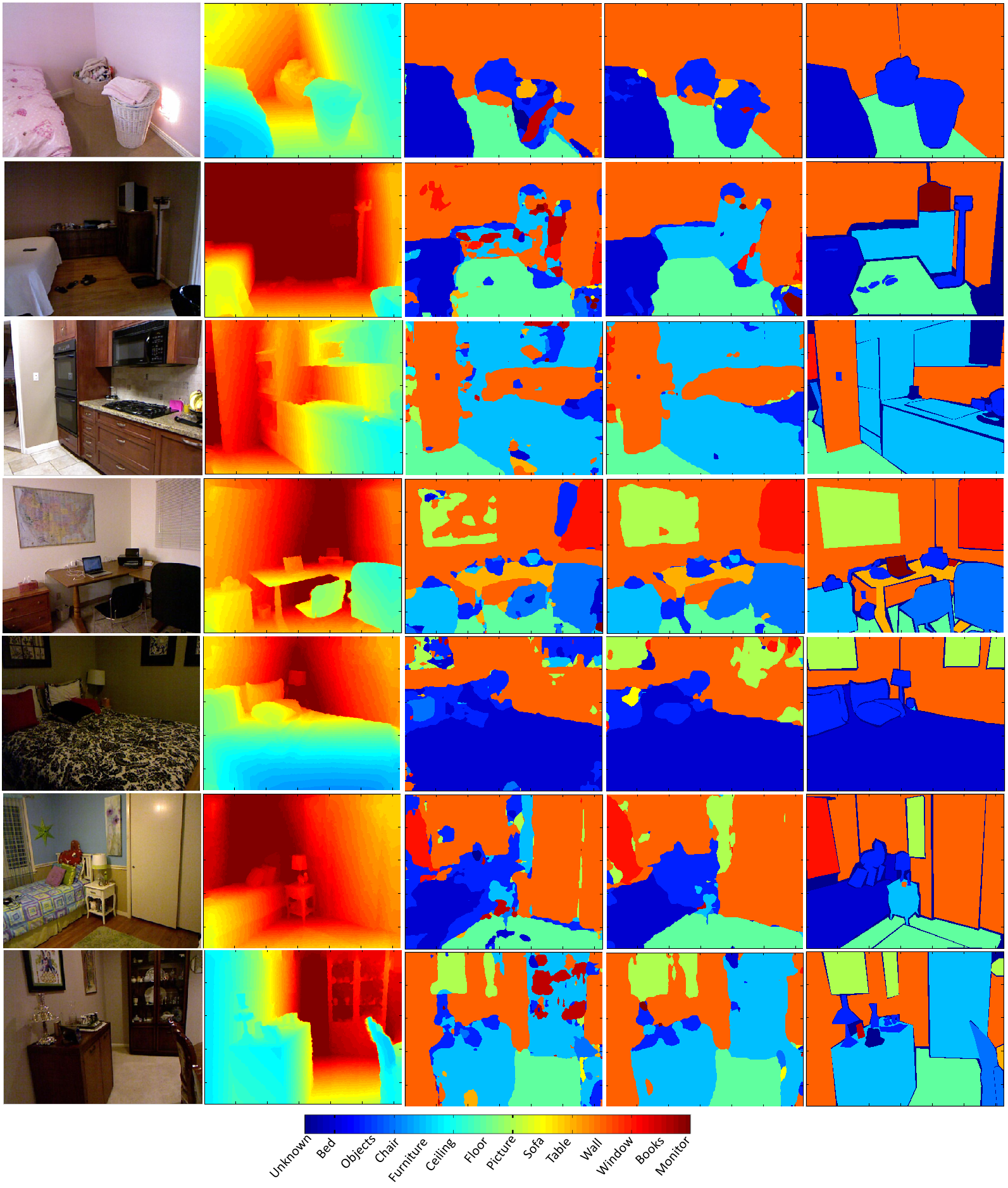}
	\end{center}
	\caption {$7$ example results from NYU-V2 (14 semantic categories). Columns 1st and 2nd: RGB color images and their corresponding Depth images. Column 3rd: the prediction results of our most competitive baseline `RNN-Features-Combined' on the NYU V2 four semantic categories. Column 4th: our multimodal-RNNs prediction results. Column 5th:ground truth. Our model is able to correctly classify many miss-classified pixels compared with our baseline. (Best viewed in color)}
	\label{missClassification3}
\end{figure*}

\begin{figure*}
	\begin{center}
		\includegraphics[width=0.7\linewidth]{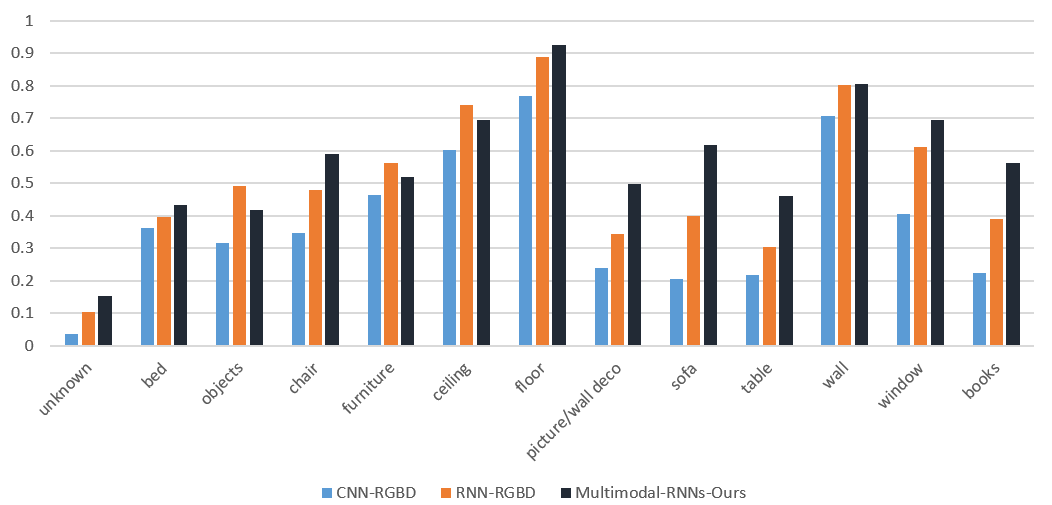}
	\end{center}
	\caption{Per class accuracy comparison between our multimodal RNNs, baselines CNN-RGBD and RNN-RGBD on NYU V2 dataset ~\cite{Silberman:indoorSegmentationv2}(14 semantic categories).}
	\label{chart14}
\end{figure*}

Figure~\ref{missClassification3} also shows some qualitative results generated by our method and the most competitive baseline method `RNN-Features-Combined'. Our multimodal-RNNs model can correctly classify many mis-classifications results compared to the baseline in most cases. We also show our per class accuracy on this sitting, as shown in Figure~\ref{chart14}. We can notice that the improvement gain of our multimodal RNNs over CNN-RGBD and RNN-RGBD models is significant. This is an evidence that our model can effectively learn powerful context-aware and multimodel features.

\begin{figure}
	\begin{center}
		\includegraphics[width=0.7\linewidth]{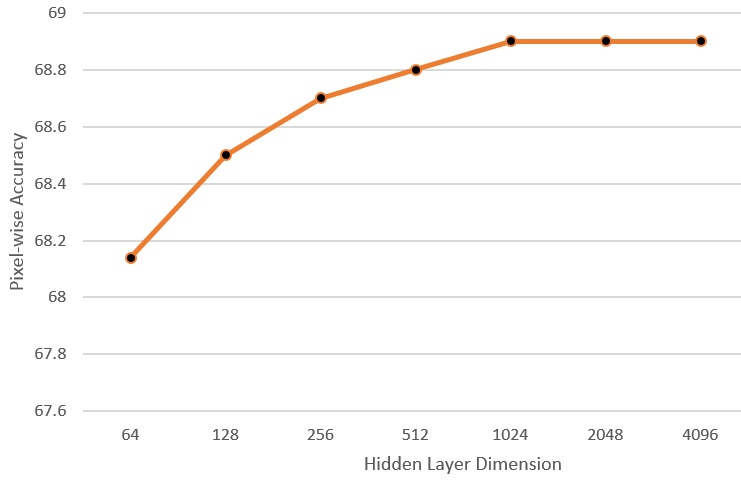}
	\end{center}
	\caption{The relationship between the hidden layer dimensions and the  pixel-wise accuracy in a single RNN model trained on the NYU V2 dataset ~\cite{Silberman:indoorSegmentationv2}}
	\label{hiddenPAramm}
\end{figure}

We also study the effect of increasing the dimensionality of the internal hidden layer on single RNN performance. We notice that RNN performs almost the same but slightly better when the hidden layer dimension increases, while it becomes extremely slow and takes a lot of time to converge. Thus, we choose the dimension of the hidden layer to be $64$. All RNN models in our baselines can achieve good performance and can converge in reasonable amount of time. Figure~\ref{hiddenPAramm} shows the relationship between a single RNN hidden layer dimensionality versus the accuracy (in terms of global pixel-wise) trained on the NYU V2 to predict $4$ semantic categories.

\begin{table}
	\begin{center}
		\begin{tabular}{|l|c|c|c|}
			\hline
			Algorithm & Pixel Acc & Class Acc & IOU \\
			\hline\hline
			CNN-RGB & 50.84\% & 35.37\% & 21.72\% \\
			CNN-Depth & 56.08\% & 36.82\% & 23.41\%\\
			CNN-RGBD & 54.40\% & 35.48\% & 22.07\%\\
			\hline
			RNN-RGB & 56.22\% & 42.41\% & 29.19\%\\
			RNN-Depth & 62.27\% & 45.99\% & 33.62\%\\
			RNN-RGBD & 61.95\% & 48.15\% & 35.74\%\\
			RNN-Features-Combined & 64.30\% & 51.54\% & 39.31\%\\
			RNN-Hiddens-Combined  & 64.50\% & 51.70\% & 37.30\%\\
			RNN-Classifiers-Combined & 64.44\% & 50.27\% & 37.42\%\\
			\hline
			Multimodal-RNNs-Ours & 66.23\% & 53.06\% & 40.59\%\\
			Multimodal-RNNs-Ours-Multiscale & \textbf{67.90}\% & \textbf{54.67}\% & \textbf{43.27}\%\\
			\hline
			Wang et al.\cite{WangAnranUnsupervised} & - & 42.20\% & -\\
			Couprie et al. \cite{indoorSemanticYannLecun}  & 52.40\% & 36.20\% & -\\
			Hermans et al. \cite{hermansMapping} & 54.20\% &  48.00\% & -\\
			Khan et al. \cite{salmanKhanGeometry}   & 58.30\% & 45.10\% & -\\
			\hline
		\end{tabular}
	\end{center}
	\caption{Results on \textbf{NYU V2} dataset \cite{Silberman:indoorSegmentationv2}. The performance of all baseline and other state-of-the-art methods is measured on total pixel-wise, average class-wise and average IOU (intersection over union) among \textbf{$14$} semantic classes. The higher the better.}
	\label{TableV2-14}
\end{table}

\textbf{Examining our multimodal RNNs with other CNN features - VGG Features on NYU V2 - 14}
We also examine our proposed multimodal RNNs while replacing the input CNN features by the extracted features from the VGG-16 pre-trained model (extracted from Conv5-3 layer) \cite{SimonyanVGG}. We focused mainly on the most competitive task among our addressed tasks, i.e. classifying the 14 classes in NYU V2. The purpose of these experiments is to validate whether our fusion structure is a network-independent model, and concretely orthogonal to other CNN networks. In other words, replacing the CNN models with more powerful network like VGG\cite{SimonyanVGG}, ResNet\cite{resnets}, FCNs\cite{fcns}, Dilated Networks\cite{dilated} and others can boost the overall performance while the relative improvement of our proposed cross-connectivity fusion is maintained. Throughout all of our experiments, we observe that \textbf{replacing our CNN features with VGG features result in a constant overall increase in the accuracies in all of our RNN models (including the baselines) of around 5\% higher in terms of IOU (most competitive metric)}.

Notice that in this paper, we didn't perform joint training between the CNN layers and our multimodal RNN layers. In contrast, we perform stage training; we first train the CNN for feature extraction and then train our RNN model for final local classification. This makes the performance comparisons on the NYU-V2 tasks between our model and other state-of-the-art models not fair (many works perform end-to-end joint training of various CNN or multi-scale CNN and CRF-based methods and even with RNN/LSTM). Notice also that end-to-end training with CNN can allow training of efficient deconvolution layers to upsample the output feature maps in order to restore their original resolution, while we use simple bilinear interpolation to upsample our final output map produced by the RNNs. Thus, in order to fairly examine the full performance of our multimodal RNN model combined with other types of recent CNN models like ResNets\cite{resnets}, FCNs\cite{fcns} and Dilated Networks\cite{dilated}; a joint end-to-end training is required. In this paper, we didn't perform this joint training between the CNN and the RNN models as it is not our main contribution, but we consider it as a very good future work. 

\subsection{Observations}
From Table \ref{TableV1}, \ref{TableV2-14} and \ref{TableV2-4} we have the following observations:

\textbf{Modeling contextual dependencies between patches using RNNs helps:} The improvement gain achieved by our RNN models over the CNN models is significant. CNN features are locally learned when performing the convolutions and thus fail to encode long-range contextual information. RNNs are powerful on modeling short and long range dependencies between patches within the image and can learn context-aware features effectively.

\textbf{Sharing information between RNNs helps:} We design the baseline `RNN-Classifiers-Combined' that combines two RNN models on the classifier level. Our multimodal RNNs model outperforms this baseline as it benefits from the shared contextual information extracted through the transfer layers.

\textbf{Learning transfer layers to connect RNNs helps:} The baselines `RNN-RGBD' and `RNN-Features-Combined' are designed to mix RGB and Depth data modalities together before learning the features. Our model outperforms these baselines because it has an assigned single RNN for each modality to retain the modality-specific information. Plus, the transfer layers are learned to adaptively extracts only the relevant multimodal shared information.
\section{Conclusion}

This paper presents a new method for RGB-D scene semantic segmentation. We introduce information transfer layers between two quad-directional 2D-RNNs. Transfer layers extract relevant contextual information across the modalities and help each modality to learn context-aware features that can capture shared information. In our future work, we will evaluate the effectiveness and the scalability of the transfer layers on more modalities (modalities $>2$).

Copyright (c) 2013 IEEE. Personal use of this material is permitted. However, permission to use this material for any other purposes must be obtained from the IEEE by sending a request to pubs-permissions@ieee.org.

\section*{Acknowledgment}
The authors would like to thank NVIDIA Corporation for their donation of Tesla K40 GPUs used in this research at the Rapid-Rich Object Search Lab. This research was carried out at both the Advanced Digital Sciences Center (ADSC), Illinois at Singapore Pt Ltd, Singapore, and at the Rapid-Rich Object Search (ROSE) Lab at the Nanyang Technological University, Singapore. This work is supported by the research grant for ADSC from A*STAR. The ROSE Lab is supported by the National Research Foundation, Singapore, under its Interactive \& Digital Media (IDM) Strategic Research Programme.

{\small
	\bibliographystyle{abbrv}
	\bibliography{egbib_cat1} 
}

\end{document}